\documentclass[runningheads]{llncs}
\usepackage{graphicx}
\usepackage{comment}
\usepackage{amsmath,amssymb} 
\usepackage{xcolor}

\usepackage{makecell}
\usepackage{multirow}
\usepackage{siunitx}
\usepackage{hyperref}
\usepackage{subfig}

\makeatletter
\newcommand{\printfnsymbol}[1]{%
  \textsuperscript{\@fnsymbol{#1}}%
}
\makeatother

\usepackage{colortbl}  
\definecolor{gray}{RGB}{87, 87, 87}
\definecolor{red}{RGB}{173, 35, 35}
\definecolor{blue}{RGB}{42, 75, 215}
\definecolor{green}{RGB}{29, 105, 20}
\definecolor{brown}{RGB}{129, 74, 25}
\definecolor{purple}{RGB}{129, 38, 192}
\definecolor{cyan}{RGB}{41, 208, 208}
\definecolor{yellow}{RGB}{189, 167, 0}
\definecolor{Red}{rgb}{0.68, 0.05, 0.0}
\definecolor{Blue}{rgb}{0.0, 0.0, 0.61}
\definecolor{Blue1}{RGB}{214, 235, 245}
\definecolor{Blue2}{RGB}{235, 245, 250}
\definecolor{lime}{RGB}{60,179,113}
\definecolor{peach}{RGB}{255, 242, 230}
\definecolor{dark_pink}{HTML}{fa438f}

\definecolor{LightBlue}{rgb}{0.88,1,1}
\definecolor{LightGreen}{HTML}{d4fcd8}
\definecolor{LightPink}{HTML}{ffe3f0}

\definecolor{JieGreen}{HTML}{d4fcd8}
\definecolor{JieYellow}{HTML}{ffffcc}
\definecolor{JieRed}{HTML}{ffeded}

\def\etal{\emph{et al.}}

\begin{document}
\pagestyle{headings}
\mainmatter
\def\ECCVSubNumber{4977}  

\title{Role-Wise Data Augmentation for\\ Knowledge Distillation} 


\titlerunning{Abbreviated paper title}
%
\author{Jie Fu\inst{1}\thanks{Equal contribution.} \and
Xue Geng\inst{2}\printfnsymbol{1} \and
Zhijian Duan\inst{3} \and
Bohan Zhuang\inst{4} \and
Xingdi Yuan\inst{5} \and \\
Adam Trischler\inst{5} \and
Jie Lin\inst{2} \and
Chris Pal\inst{1} \and
Hao Dong\inst{3}
}
%
%
\institute{Mila, Polytechnique Montr\'eal, \email{jie.fu@polymtl.ca}\and
I$^2$R, A*STAR
\\ \and
Peking University\\ \and
Monash University \and 
Microsoft Research Montr\'eal
}
\maketitle

\begin{abstract}
Knowledge Distillation (KD) is a common method for transferring the ``knowledge'' learned by one machine learning model (the \textit{teacher}) into another model (the \textit{student}),
where typically, the teacher has a greater capacity (e.g., more parameters or higher bit-widths). 
To our knowledge, existing methods overlook the fact that although the student absorbs extra knowledge from the teacher, both models share the same input data -- and this data is the only medium by which the teacher's knowledge can be demonstrated. 
Due to the difference in model capacities, the student may not benefit fully from the same data points on which the teacher is trained. 
On the other hand, a human teacher may demonstrate a piece of knowledge with individualized examples adapted to a particular student, for instance, in terms of her cultural background and interests.
Inspired by this behavior, we design data augmentation agents with distinct roles to facilitate knowledge distillation.
Our data augmentation agents generate distinct training data for the teacher and student, respectively.
We find empirically that specially tailored data points enable the teacher's knowledge to be demonstrated more effectively to the student.
We compare our approach with existing KD methods on training popular neural architectures and demonstrate that role-wise data augmentation improves the effectiveness of KD over strong prior approaches.
The code for reproducing our results can be found at \url{https://github.com/bigaidream-projects/role-kd}
\end{abstract}

\section{Introduction}


In the educational psychology literature, it is generally considered beneficial if teachers can adapt curricula based upon students' prior experiences \cite{bandura2002cultural,brumfiel2005has,gurlitt2006can,slotta2006helping}.
These vary widely depending on students' cultural backgrounds, previous educational experiences, interests, and motivations.

Knowledge distillation (KD) \cite{Bucilua2006,Hinton2014} is a common framework for training machine learning models. It works by transferring knowledge from a higher-capacity teacher model to a lower-capacity student model. 
Most KD methods can be categorized by how they define the knowledge stored in the teacher (i.e., the ``soft targets'' of training as defined in existing literature).
For instance, \cite{Hinton2014} originally proposed KD for neural networks, and they define the output class probabilities (i.e., soft labels) generated by the teacher as the targets for assisting the training of students. 
In a follow up work, \cite{Romero2014} defined the soft targets via the feature maps in the teacher model's hidden layers. 

To train a student network with KD effectively, it is important to distill as much knowledge from the teacher as possible. 
However, previous methods overlook the importance of the \textit{medium} by which the teacher's knowledge is demonstrated: the training data points. 
We conjecture that there exist examples, not necessarily seen and ingested by the teacher, that might make it easier for the student to absorb the teacher's knowledge.
Blindly adding more training examples may not be beneficial because it may slow down training and introduce unnecessary biases \cite{Ho2019}.
The analogy with how human teachers adjust their teaching to their students' particular situations (e.g., with the feedback gathered from the students during teaching) suggests that a reasonable yet uninvestigated approach might be to augment the training data for both the teacher and student according to \textit{distinct} policies.

In this paper, we study whether and how adaptive data augmentation and knowledge distillation can be leveraged synergistically to better train student networks.
We propose a two-stage, role-wise data augmentation process for KD. This process consists of: (1) training a teacher network till convergence while learning a schedule of policies to augment the training data specifically for the teacher; (2) distilling the knowledge from the teacher into a student network while learning another schedule of policies to augment the training data specifically for the student.  
It is worth noting that this two-stage framework is orthogonal to existing methods for KD, which focus on how the knowledge to be distilled is defined; thus, our approach can be combined with previous methods straightforwardly.


Although our proposed method can in principle be applied to any models trained via KD, we focus specifically on how to use it to transfer the knowledge from a full-precision teacher network into a student network with lower bit-width.
Network quantization is crucial when deploying trained models on embedded devices, or in data centers to reduce energy consumption \cite{Strubell2019}. 
KD-based quantization \cite{Zhuang2018,Polino2018} jointly trains a full-precision model, which acts as the teacher, alongside a low-precision model, which acts as the student.
Previous work has shown that distilling a full-precision teacher's knowledge into a low-precision student, followed by fine-tuning, incurs noticeable performance degradation, especially when the bit-widths are below four \cite{Zhuang2018,Polino2018}. 
We show that it is advantageous to use adaptive data augmentation to generate more training data for the low-precision network based on its specific weaknesses. 
For example, low-precision networks may have difficulties learning rotation-related patterns,\footnote{We will visualize the learned schedules of policies in Section \ref{sec:visualize}.} and the data augmentation agent should be aware of this and generate more such data points.
\section{Related Work}

\paragraph{Knowledge distillation.}

KD is initially proposed for model compression, where a powerful wide/deep teacher distills knowledge to a narrow/shallow student to improve her performance~\cite{Hinton2014,Romero2014}. 
Most KD methods mainly differ in how they define the knowledge learned by the teacher. 
For example, \cite{Hinton2014} define the class probabilities (i.e., soft labels) generated by the teacher as the knowledge, and \cite{Romero2014} treat the teacher's feature maps as the knowledge to be transferred. 
Moreover, some literature~\cite{Zhang_2018_CVPR,Park_2019_CVPR,GraphKD,Lee_2018_ECCV} designs advanced distillation strategies in order to let the student learn better from the teacher's rich knowledge.
Due to the effectiveness of KD, it has also been widely used in many computer vision tasks. For example, Zhang~\etal~\cite{zhang2016real} propose to transfer the knowledge learned with optical flow CNN to improve the action recognition performance. 
And several works propose to learn efficient object detection~\cite{chen2017learning,wei2018quantization} and semantic segmentation~\cite{he2019knowledge} with distillation. 
In terms of the definition of knowledge to be distilled from the teacher, existing models typically use teacher's class probabilities~\cite{Hinton2014} and/or intermediate features~\cite{Romero2014,Park_2019_CVPR,GraphKD,Lee_2018_ECCV}. 
Among those KD methods that utilize intermediate feature maps, Relational KD (RKD) considers \cite{Park_2019_CVPR} the intra-relationship in the same feature map, while Multi-Head KD (MHKD)\cite{GraphKD} and KD using SVD (KD-SVD) \cite{Lee_2018_ECCV} utilize the inter-relationship across feature maps. 
Compared to these relationship-based KD methods, we incorporate both the intra- and inter-relationships within and across feature maps, which is an additional engineering trick used in our paper. 
More importantly, all the previous KD methods ignore the dual-role of the training data for the teacher and the student. 
In this paper, we propose to use different training data for the teacher and the student, where the training data is augmented by distinct learned policies. 

\paragraph{Automated data augmentation.}

Manually applying data augmentation rules such as random rotating, flipping, and scaling are common practices for training neural models on image classification tasks \cite{Krizhevsky2012,He2016}. 
Several recent works attempt to automate the data augmentation process. 
Generative adversarial networks \cite{Ratner2017} and Bayesian optimization \cite{Tran2017} have been used for this process. 
\cite{DeVries2017} augment training data in the learned feature space by injecting noise and interpolation. 
\cite{Lemley2017} learn how to combine pairs of images for data augmentation. 
AutoAugment \cite{Cubuk2018} searches for the optimal data augmentation policies (e.g., how to rotate) based on reinforcement learning. 
However, the search process is computationally expensive. 
Population-based augmentation (PBA) \cite{Ho2019} uses an evolution-based algorithm to automatically augment data in an efficient way. 
In contrast to previous approaches, we study the effect of the training data for different roles in KD (i.e., the teacher and the student) and propose to use automatic data augmentation to train the student better from her teacher.

\paragraph{Network quantization.} Quantization is an effective approach for compressing models by using low bitwidth weights and activations. It can be categorized into fixed-point quantization and binary neural networks. 
Uniform approaches~\cite{Zhou2016,Zhuang2018} perform fixed-point quantization with a constant quantization step. To reduce the quantization error, non-uniform strategies~\cite{Zhang2018,Jung2018,Cai2017} propose to jointly learn the quantizer and model parameters for better accuracy. 
Moreover, to relax the non-differentiable quantizer, which is core issue of quantization, some works propose to make the gradient-based optimization feasible by using gumble softmax~\cite{Louizos2019} or learning with regularization~\cite{Bai2019}. 
KD methods have shown to be effective at improving the performance of lower-capacity networks using the knowledge from higher-capacity networks \cite{Mishra2018,Zhuang2018}. 
However, the KD methods used in \cite{Mishra2018,Zhuang2018} are generic and not tailored to the quantization problems. 
In this paper, our proposed KD method takes into account the fact that the student may prefer different training data from the teacher. 












\section{Preliminaries}

\subsection{Population-Based Augmentation (PBA)}
Population Based Augmentation (PBA) \cite{Ho2019} is an algorithm that quickly and efficiently learns data augmentation functions for neural network training. 
Instead of generating a fixed augmentation policy, as an evolutionary search algorithm, PBA learns a dynamic \textit{per-epoch} schedule of augmentation policies, denoted as $\mathcal{A}$. Since this schedule is epoch-based, it will \textit{re-create} the augmented dataset every epoch.
PBA begins with a population of models that are trained in parallel on a small subset of the original training data. 
The weights of the worse performing models in the population are replaced by those from better performing models (i.e., exploitation), and the augmentation policies $\mathcal{A}$ are changed to new ones from the pre-defined policy search space (i.e., exploration). 
More concretely, $\mathcal{A}$ consists of a series of vectors of $(operator, probability, magnitude)$, and PBA learns a schedule to get a new datapoint $\tilde{x_i}=operator(x_i,magnitude)$ with an adaptive $probability$. 
Following \cite{Ho2019}, we use 15 common operators such as random cropping, flipping, scaling, rotating, and translating. 
Note that when $probability=0$, it is equivalent to a null operator. 

After training, PBA usually keeps the learned augmentation schedule of policies but \textit{discards} the elementary parameters of the models. The discovered schedules of the small subset can be used directly on the original training data. Even more, a different model (e.g., larger one) can also use the learned schedule to improve their training on the same task.



\subsection{Knowledge Distillation (KD)\label{sec:method_kd}}

Following the notations in \cite{Park_2019_CVPR}, a KD method aims to minimize the objective function:
\begin{equation}
    \mathcal{L}_\text{general} = \mathcal{L}_{\text{task}} + \lambda \cdot \mathcal{L}_{\text{KD}},
\end{equation}
where $\lambda$ is a hyper-parameter to balance the impact of the KD loss term.

In this paper for classification tasks,
\begin{equation}
\mathcal{L}_{\text{task}} = \sum_{x_i \in \mathcal{X}} \mathcal{H}(\text{softmax}(\mathcal{F}_S^{\text{final}}(x_i)) , y_{\text{truth}}),     
\label{eq:task}
\end{equation}
where $\mathcal{X}$ refers to training sample space, $y_{\text{truth}} \in \mathcal{Y}$ are the ground-truth labels, $\mathcal{F}_S(\cdot)$ is the student network, and $\mathcal{H}(\cdot)$ denotes the cross-entropy.

The KD term can be defined as:
\begin{equation}
    \mathcal{L}_{\text{KD}} = \sum_{x_i \in \mathcal{X}} l(\mathcal{F}_T(x_i), \mathcal{F}_S(x_i)),
\end{equation}
where $\mathcal{F}(\cdot)$ is the function of the network.
And $l(\cdot)$ is a loss function to compute the difference between the teacher network and the student network.

For KD methods \cite{Hinton2014} that use soft labels, the objective can be defined as:

\begin{equation}
    \mathcal{L}_{\text{KD}}^{\text{soft}} = \sum_{x_i \in \mathcal{X}} \mathcal{H}(\text{softmax}(\mathcal{F}_T^{\text{final}}(x_i)) , \text{softmax}(\mathcal{F}_S^{\text{final}}(x_i))),
\end{equation}
where $\mathcal{F}^{\text{final}}(x_i)$ is the feature map of the final layer.


We notice that there exists some KD methods utilizing the intermediate feature maps in complementary ways. 
For example, Relational KD \cite{Park_2019_CVPR} considers the intra-relationships. 
That is, given the feature map of layer $j$, the KD loss can be formulated as:
\begin{equation}
    \mathcal{L}_{\text{KD}}^{\text{intra}} = \sum_{x_i \in \mathcal{X}} l(\Phi (\mathcal{F}^j_T(x_i)), \Phi (\mathcal{F}^j_S(x_i))),
\end{equation}
where $\Phi(\cdot)$ refers to the potential function measuring the pairwise relationship inside a feature map from student network or teacher network and $\mathcal{F}^j(x_i)$ is the feature map of layer $j$, including the final logits layer. 
Therefore, this feature-based KD method includes the benefits of using soft labels. 

On the other hand, some works \cite{GraphKD,Lee_2018_ECCV} consider the inter-relationships, where the KD term can be formulated as:
\begin{equation}
    \mathcal{L}_{\text{KD}}^{\text{inter}} = \sum_{x_i \in \mathcal{X}} l(\varphi (\mathcal{F}^j_T(x_i), \mathcal{F}^k_T(x_i)), \varphi (\mathcal{F}^j_S(x_i), \mathcal{F}^k_S(x_i))),
\end{equation}
where $\varphi(\cdot)$ measures the inter-relationship between feature maps of different layers, i.e. $k \neq j$. 

\subsection{Quantization}
\label{sec:quantize}
In this work, we use DoReFa\footnote{It should be noted that our proposed method is orthogonal to any particular quantization method.} \cite{Zhou2016} to quantize both weights and activations. 
The quantization function $Q(\cdot)$ is defined as:
\begin{equation}
    r_q = Q(r) = \frac{1}{2^{n_{\text{bits}}}-1} \cdot \text{round}((2^{n_{\text{bits}}}-1) \cdot r),
\label{eq:quantization1}
\end{equation}
where $r$ is the full-precision value, $r_q$ indicates the quantized value, $n_{\text{bits}}$ refers to the number of bits to represent this value. With this quantization function, the quantization on weights $w$ is defined as:
\begin{equation}
    w_q = 2 \cdot Q(\frac{\text{tanh}(r)}{2 \cdot \text{max}(|\text{tanh}(w)|)} + \frac{1}{2}) - 1.
    \label{eq:quantization1_w}
\end{equation}

The back-propagation is approximated by the straight-through estimator \cite{Bengio2013}, the partial gradient $\frac{\partial l}{ \partial r}$ w.r.t. the loss $l$ is computed as:
\begin{equation}
\frac{\partial l}{ \partial r} = \frac{\partial l}{ \partial r_q} \cdot \frac{\partial r_q}{ \partial r} \approx \frac{\partial l}{ \partial r_q}.
    \label{eq:ste}
\end{equation}

\section{The Proposed Method}
Our proposed method has two stages which will be described in the following subsections. 
In the first stage (Stage-$\alpha$), we train a teacher network, denoted as $\mathcal{N}_T$, with the help of PBA-based augmentation.
In the second stage (Stage-$\beta$), we further distill the knowledge from $\mathcal{N}_T$ (pre-trained in the first stage) to the student network, denoted as $\mathcal{N}_S$, while learning another augmentation schedule to augment the training data for $\mathcal{N}_S$. 


\subsection{Stage-$\alpha$}\label{sec:stage_one}
\vspace{-1em}
\begin{figure}
\centering\includegraphics[width=0.8\textwidth]{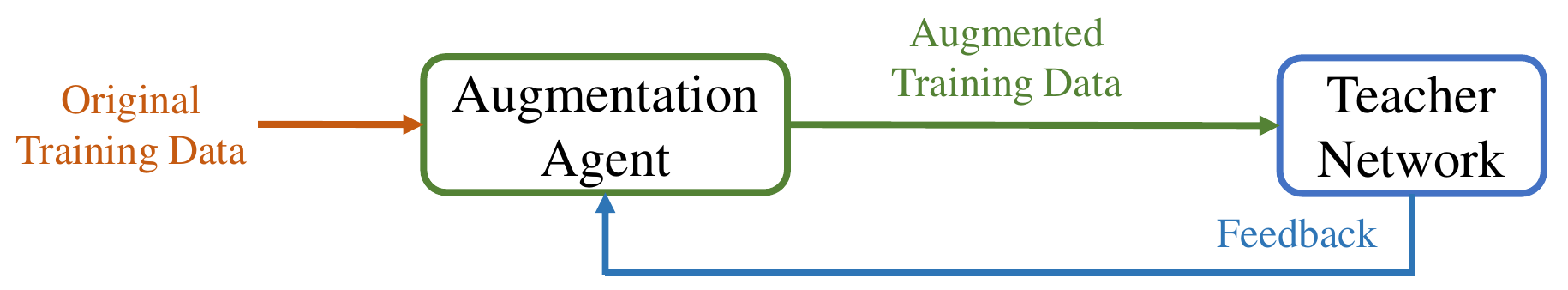}
\caption{Diagram (stage-$\alpha$) of training a data augmentation agent for the $\mathcal{N}_T$. 
\label{fig:aug}}
\end{figure}

In general, a teacher can provide better training signals for the student if the teacher's performance increases \cite{Mirzadeh2019}. 
As shown in Fig. \ref{fig:aug}, we apply PBA to learn an dynamic per-epoch schedule of augmentation policies, $\mathcal{A}_T$, for $\mathcal{N}_T$ on a small \textit{subset} of training data. 
That is, the augmentation agent's training signal is defined as the feedback of $\mathcal{N}_T$'s accuracy on a subset of the dataset. 
After this, we use the discovered schedule $\mathcal{A}_T$ to augment the whole training dataset and re-train $\mathcal{N}_T$ on it till convergence.

Stage-$\alpha$ can be seen as a preparation stage which is used to improve the teacher's performance and does not interfere with the augmentation policy learning for the student. 
This can been also shown in Fig. \ref{fig:aug}, where the feedback is only from $\mathcal{N}_T$.

\subsection{Stage-$\beta$}
\label{sec:stage_two}



\begin{figure}
\centering\includegraphics[width=0.90\textwidth]{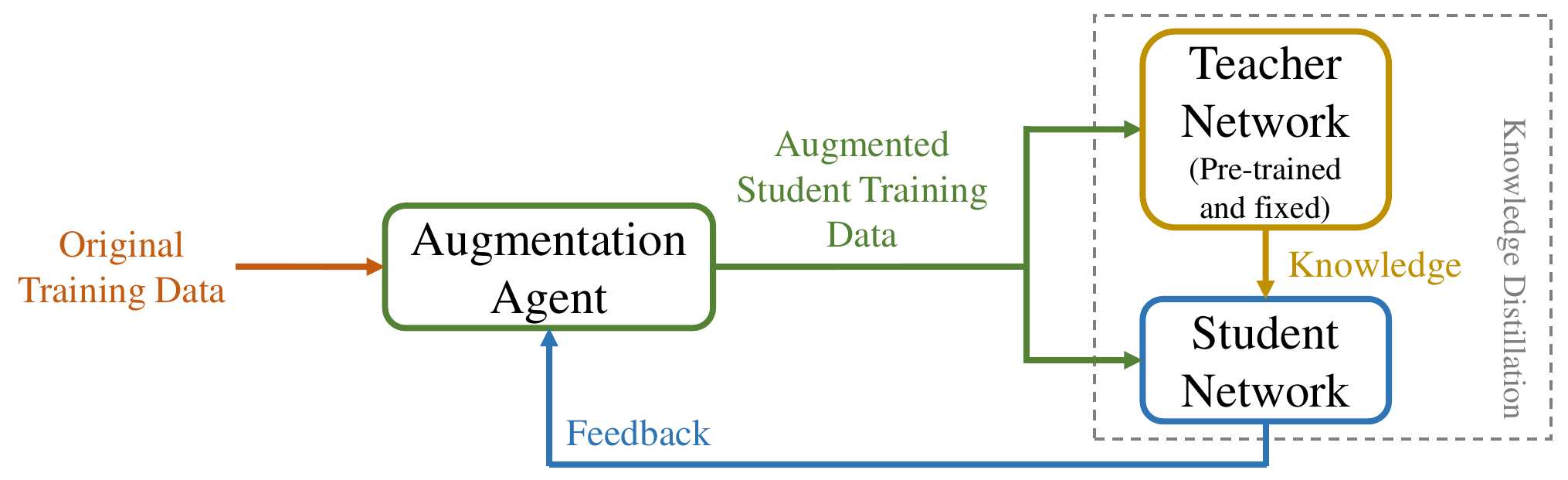}
\caption{Diagram (stage-$\beta$) to augment training datapoints for both $\mathcal{N}_T$ and $\mathcal{N}_S$. 
The $\mathcal{N}_T$ has been pre-trained using the method shown in Section \ref{sec:stage_one}, and is fixed during training. 
The augmentation agent in stage-$\beta$ is designed to learn schedules of polices that are different from those learned in stage-$\alpha$, and thus the agent only receives the feedback from $\mathcal{N}_S$. 
\label{fig:quan_kd}}
\end{figure}

In this stage, we distill the knowledge from $\mathcal{N}_T$ (pre-trained in Stage-$\alpha$) to the student network, denoted as $\mathcal{N}_S$, while learning another augmentation schedule to augment the training data for $\mathcal{N}_S$. 
In order to take advantage of this functionality, we apply the KD methods together with data augmentation in stage-$\beta$ as shown in Fig. \ref{fig:quan_kd}.

More concretely, we first use PBA to learn an epoch-based augmentation schedule $\mathcal{A}_S$ for $\mathcal{N}_S$ on a subset of the dataset. 
Different from the schedule $\mathcal{A}_T$ learned in stage-$\alpha$, $\mathcal{A}_S$ is learned based on the feedback (i.e., accuracy) from $\mathcal{N}_S$, who is trained with KD. 
In other words, $\mathcal{N}_S$ receives additional training signals from $\mathcal{N}_T$ that is pre-trained in stage-$\alpha$. 
The augmentation policy $\mathcal{A}_S$ is learned to facilitate this KD process, and thus would be different from $\mathcal{A}_T$ that is used by the teacher $\mathcal{N}_T$. 

It should be noted that the above process is only used to learn the augmentation policy $\mathcal{A}_S$ for the student on a subset of the whole training data. 
After this, we use the learned $\mathcal{A}_S$ to augment the whole training dataset, and \textit{re-train} $\mathcal{N}_S$ on it with the distilled knowledge from $\mathcal{N}_T$.
Note that, because the learned schedule is epoch-based, we do not use the discovered schedule $\mathcal{A}_T$ from stage-$\alpha$ to augment the training data as initialization. 

When $\mathcal{N}_S$ is a low-precision network, following \cite{Furlanello2018}, we share the same network architecture\footnote{Note this is not a hard constraint, we choose such strategy to reduce the number of factors that might influence the final performance.} between $\mathcal{N}_T$ and $\mathcal{N}_S$. 
When $\mathcal{N}_S$ is a full-precision network, she will have fewer layers compared to $\mathcal{N}_T$. 




\vspace{-1em}

\section{Experiments}

\subsection{Settings}

We evaluate our approach on two benchmark datasets: CIFAR-10 \cite{Krizhevsky2009} and CIFAR-100. 
We search over a ``reduced'' CIFAR-10/CIFAR-100 with 4,000 training images and 36,000 validation images, which is the same as in \cite{Ho2019}. 
All the data augmentation models are run with 16 total trials to generate augmentation schedules. Following PBA, in stage-$\alpha$, we run PBA to create schedules over separate models and then transfer the CIFAR-10 policy to CIFAR-100. However, for student network training in stage-$\beta$, we empirically use the respective ``reduced'' dataset.
The data augmentation approaches for the baselines include random crop and horizontal flipping operations.
Following \cite{Ho2019}, our policy search space has a total of 15 operations, each having two magnitude and discrete probability values. 
We use discrete probability values from 0\% to 100\%, in increments of 10\%. 
Magnitudes range from 0 to 9. 


For the verification on quantization, the models we evaluate on include AlexNet \cite{Krizhevsky2012} and ResNet18 \cite{He2016}. For the verification on full-precision networks in Section \ref{sec:full_precision_experiment}, the networks we evaluate are Wide Residual Network \cite{BMVC2016_87}, PyramidNet \cite{DPRN} and PreResNet \cite{he2016identity}. 

The number of epochs is 200 and the batch size is 128. For full-precision network $\mathcal{N}_T$, the learning rate starts from 0.1 and is decayed by 0.1 after every $30\%$ of the total epochs. we use SGD with a Nesterov momentum optimizer. 
The weight decay is set to $5\times 10^{-4}$. 
For quantization, the learning rate is set to $10^{-3}$ and is divided by 10 every $30\%$ of the total epochs. 
We use the pre-trained teacher network model as the initial point of student network.
We use a smaller weight decay $10^{-5}$ assuming that less regularization is needed by the lower-precision networks. Following DoReFa \cite{Zhou2016}, the first layer and last layer are not quantized. 

Following \cite{Clark2019}, during training, we gradually transition the student from learning based on the teacher to training based on the ground-truth labels.
This heuristic provides the student with more rich training signals in the early stage but does not force the student to strictly mimic the teacher's behaviors. 
As for the implementation, we decay the balancing hyper-parameter $\lambda$ in the KD loss by 0.5 every 60 epochs.

\subsection{Comparing Different KD Methods\label{sec:kd_experiment}}

As mentioned in Section \ref{sec:method_kd}, there exist complementary KD methods considering both intra- and inter-relationships within and across feature maps. 
A natural question is if it would be beneficial to combine them to further boost the performance together with data augmentation. 
Therefore, we propose a simple extension to these complementary KD methods, dubbed as II-KD, by incorporating intra-relationships inside the feature map and inter-relationships across different feature maps. 
We incorporate the two relationships into the final objective function as follows:
\begin{equation}
    \mathcal{L}_{\text{KD}}^{\text{II}} = \mathcal{L}_{\text{task}} + \lambda \cdot (\mathcal{L}_{\text{KD}}^{\text{intra}} + \mathcal{L}_{\text{KD}}^{\text{inter}}), 
\label{eq:total_loss}
\end{equation}
where we only use a single balancing hyper-parameter $\lambda$ between the original loss and the distillation loss, which does not introduce extra hyper-parameters.


More precisely, our KD method incorporates components of three conventional KD methods: RKD \cite{Park_2019_CVPR}, MHGD \protect\cite{GraphKD} and KD-SVD \protect\cite{Lee_2018_ECCV}.
As shown in Eq. (\ref{eq:total_loss}), we add the three KD terms together with equal coefficients. 
We use the loss function $l(\cdot)$ following their approaches. For the back-propagation, we clip the gradient for KD loss as in KD-SVD, because this will smoothly post-processes the gradient to limit the impact of KD loss in training.
For AlexNet we select the feature maps of ReLU layers after the convolution/max pooling layer. 
For ResNet18, we select the feature maps of the last ReLU layer of each residual block. 

We evaluate our proposed KD extension on CIFAR-100 with ResNet18 for different bit-width settings by comparing with various KD methods. 
For the baseline methods, we use their default settings with a fixed and pre-trained teacher network in the training stage and $\lambda = 1$ for the knowledge distillation loss. 
We set $\lambda = 0.4$ for II-KD in Eq. (\ref{eq:total_loss}), as we have two KD terms. 
Tab.~\ref{tab:kd_comp} reports the results on various augmented KD methods. 
We observe that our proposed methods clearly outperform the other KD methods on all the settings, though the improvements over MHGD and KD-SVD are not huge. 
The results also reveal that only relying soft labels is not as effective as utilizing multiple supervising signals from the teacher. 
It should be noted that we do not thoroughly tune the coefficients within II-KD, as we want to minimize the number of hyperparameters and we find that it is effective with the most simple setting. 

\begin{table}[t]
\centering
\caption{Accuracy comparison among various KD methods for CIFAR-100 using ResNet18 with 2-bit/4-bit settings. We compare ours with the following methods: Soft labels \protect\cite{Hinton2014}, DML \protect\cite{Zhang_2018_CVPR}, RKD \protect\cite{Park_2019_CVPR}, MHGD \protect\cite{GraphKD} and KD-SVD \protect\cite{Lee_2018_ECCV}. }
\small
  \begin{tabular}{ c | c | c | c | c | c | c  }
  \hline
  \makecell{Bit-Width \\ (Weight / Activation)} & Soft labels & DML & RKD & MHGD & KD-SVD & II-KD \\ \hline
  
  4/4 & 70.48 & 72.47 & 71.84 & 73.52 &  73.92 & \cellcolor{JieGreen}74.21 \\ \hline
  2/2 & 70.09 & 69.72 & 70.71 & 71.80 & 72.97 & \cellcolor{JieGreen}73.35  \\ \hline
\end{tabular}
\label{tab:kd_comp}
\vspace{-5mm}
\end{table}



\subsection{Is Role-Wise Augmentation with KD Effective for Quantization?} 
\label{sec:low_precision_experiment}

In this subsection, we aim to answer this question: is our two-stage role-wise augmentation with KD effective for network quantization?
We conduct experiments on CIFAR-10 and CIFAR-100 datasets under full-precision, 4-bit, and 2-bit settings. 

From Tab.~\ref{tab:baseline_comp}, we can observe that training with learned data augmentation schedules does not improve the performance of low-precision networks too much. 
Similar to the results obtained in \cite{Zhuang2018}, transferring knowledge from the full-precision to the low-precision student usually helps the training of students, which is especially obvious on the CIFAR-100 dataset. 
Tab.~\ref{tab:baseline_comp} also clearly shows that our proposed pipeline consistently improves the performance of the low-precision student networks. 
For example, the 4-bit $\mathcal{N}_S$ is comparable with full-precision reference without loss of accuracy for CIFAR-10 and with loss of accuracy within 1.0\% on CIFAR-100. 
When decreasing the numerical precision to 2-bit, the results are still promising as compared with other baselines, even though there is a performance gap between the 2-bit and the full-precision models. 
For instance, our approach usually outperforms the strong baseline, only using II-KD, by more than $1.0\%$. In particular, we observe that when the numerical precision is lower, the augmentation probably help more in improving the performance of KD methods as comparing the performance of II\_KD and Stage-$\beta$. For instance, our proposed methods has 0.91\% gain in 2-bit but only has 0.36\% gain in 4-bit with ResNet18 network for CIFAR-100.

\begin{table}[t]
\centering
\caption{Accuracy on CIFAR-10 and CIFAR-100 datasets with different bit-widths.
\textbf{Vanilla} for 4-bit and 2-bit refers to training a network based on DoReFa \cite{Zhou2016} from scratch without learned data augmentation. 
\textbf{Stage-$\alpha$} refers to using learned schedules discovered by PBA to \textit{re-train} $\mathcal{N}_T$ as described in Section \ref{sec:stage_one}.
\textbf{only II-KD} refers to training $\mathcal{N}_S$ using II-KD but without the learned data augmentation.
\textbf{Stage-$\beta$} refers to training $\mathcal{N}_S$ using II-KD and the learned data augmentation. 
For \textbf{Vanilla} and \textbf{Stage-$\alpha$}, we report the accuracy of $\mathcal{N}_T$, and for the rest we report the accuracy of $\mathcal{N}_S$.}
\small
  \begin{tabular}{ c | c | c | c | c | c }
  \hline
  \multicolumn{2}{c|}{Methods}  & \makecell{AlexNet \\ CIFAR-10} & \makecell{AlexNet \\ CIFAR-100 } & \makecell{ResNet18 \\ CIFAR-10} & \makecell{ResNet18 \\ CIFAR-100} \\ \hline
  \multirow{3}{*}{Vanilla} 
  & \cellcolor{JieGreen}32-bit  & \cellcolor{JieGreen}90.58 & \cellcolor{JieGreen}65.80 & \cellcolor{JieGreen}93.57 & \cellcolor{JieGreen}74.85 \\ \cline{2-6} 
  & \cellcolor{JieYellow}4-bit  & \cellcolor{JieYellow}89.72 &\cellcolor{JieYellow} 60.25 & \cellcolor{JieYellow}90.97 & \cellcolor{JieYellow}69.81 \\ \cline{2-6}
  & \cellcolor{JieRed}2-bit  & \cellcolor{JieRed}88.77 & \cellcolor{JieRed}58.96 & \cellcolor{JieRed}90.00 & \cellcolor{JieRed}67.06 \\ \hline
  \multirow{3}{*}{Stage-$\alpha$} 
  & \cellcolor{JieGreen}32-bit  & \cellcolor{JieGreen}91.62 & \cellcolor{JieGreen}66.40 & \cellcolor{JieGreen}94.49 & \cellcolor{JieGreen}75.19 \\ \cline{2-6} 
  & \cellcolor{JieYellow}4-bit  & \cellcolor{JieYellow}90.06 & \cellcolor{JieYellow}60.65 & \cellcolor{JieYellow}91.47 & \cellcolor{JieYellow}70.24 \\ \cline{2-6}
  & \cellcolor{JieRed}2-bit  & \cellcolor{JieRed}89.28 & \cellcolor{JieRed}58.59 & \cellcolor{JieRed}89.99 & \cellcolor{JieRed}67.32 \\ \hline
  \multirow{2}{*}{ Only II-KD}
  & \cellcolor{JieYellow}4-bit & \cellcolor{JieYellow}90.55 &\cellcolor{JieYellow} 65.55 & \cellcolor{JieYellow}91.42 & \cellcolor{JieYellow}73.85 \\ \cline{2-6}
  & \cellcolor{JieRed}2-bit & \cellcolor{JieRed}89.18 & \cellcolor{JieRed}63.49 & \cellcolor{JieRed}90.60 & \cellcolor{JieRed}72.44 \\ \hline
  
  \multirow{2}{*}{Stage-$\beta$}
  & \cellcolor{JieYellow}4-bit & \cellcolor{JieYellow}92.00 &\cellcolor{JieYellow} 65.69 & \cellcolor{JieYellow}94.44 & \cellcolor{JieYellow}74.21 \\ \cline{2-6}
  & \cellcolor{JieRed}2-bit & \cellcolor{JieRed}90.63 & \cellcolor{JieRed}64.06 & \cellcolor{JieRed}93.20 & \cellcolor{JieRed}73.35 \\ \hline

  \end{tabular}
\label{tab:baseline_comp}
\vspace{-4mm}
\end{table}

\subsection{Comparing Schedules}
\label{sec:tch_aug_experiment}

Here we aim to answer this question: how effective is it if we use $\mathcal{A}_T$, learned based on the feedback from $\mathcal{N}_T$ in stage-$\alpha$, to dynamically augment the training dataset and train $\mathcal{N}_S$ on it.
Tab. \ref{tab:aug_comp} reports the accuracy comparison with different KD methods and augmentation schedules. 
We can clearly see that augmenting the training dataset for $\mathcal{N}_S$ with $\mathcal{A}_S$ consistently outperforms those using the transferred schedules $\mathcal{A}_T$ among different KD methods. 
This observation is consistent with our assumption that $\mathcal{N}_S$ has her own optimal augmentation schedule, $\mathcal{A}_S$, that is different from $\mathcal{A}_T$ for $\mathcal{N}_T$. 
In particular, blindly applying the teacher augmentation schedule $\mathcal{A}_T$ may negatively influence the training of $\mathcal{N}_S$ as compared to only using KD. 
For example, the learned schedule based on the teacher $\mathcal{A}_T$ degrades the performance of $\mathcal{N}_S$ by 0.58\% for AlexNet on CIFAR-100 as compared to applying KD methods, as shown in Tab. \ref{tab:baseline_comp}. Furthermore, we observe that II-KD outperforms other KD methods significantly on CIFAR-100. This observation probably proves that the combination of intra- and inter- KD methods helps to boost the classification performance.


\begin{table}[t]
\centering
\caption{Accuracy comparison on learned schedules from teacher and student separately for CIFAR-100 with 4-bit networks using different KD methods.}
\small
\begin{tabular}{c|c|c|c|c|c|c}
\hline
& \multicolumn{3}{c|}{AlexNet} 
& \multicolumn{3}{c}{ResNet18} \\ \hline
Methods & DML & MHGD & II-KD & DML & MHGD & II-KD  \\ \hline
Schedules based on teacher & 61.61 & 61.62 & 64.97 & 71.78 & 69.76 & 73.46 \\ \hline
Schedules based on student & \cellcolor{JieGreen}63.73 & \cellcolor{JieGreen}63.47 & \cellcolor{JieGreen}65.69 & \cellcolor{JieGreen}72.47 & \cellcolor{JieGreen}73.52 & \cellcolor{JieGreen}74.21 \\ \hline
\end{tabular}

\label{tab:aug_comp}
\vspace{-4mm}
\end{table}

\subsection{Analyzing the Learned Schedules\label{sec:visualize}}
\label{sec:vis_experiment}
\begin{figure}[!t]
\centering
\subfloat[Normalized plot of operation probability parameters over time for the teacher network $\mathcal{N}_T$. \label{fig:teacher_prob}]{\includegraphics[width=0.9\textwidth]{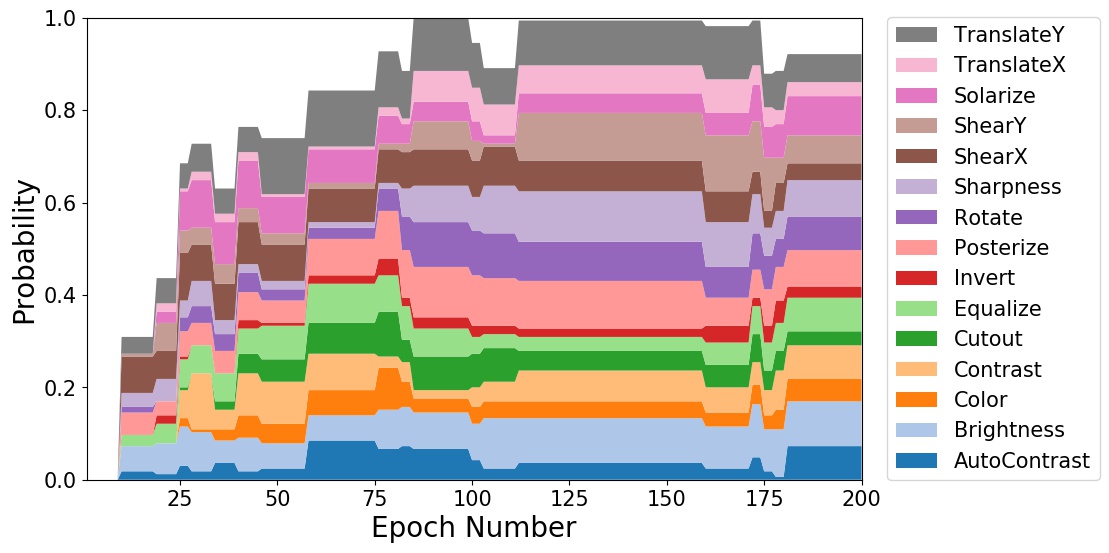}}\hfill
\subfloat[Operation magnitude parameters over time for the teacher network $\mathcal{N}_T$.\label{fig:teacher_mag}]{\includegraphics[width=0.9\textwidth]{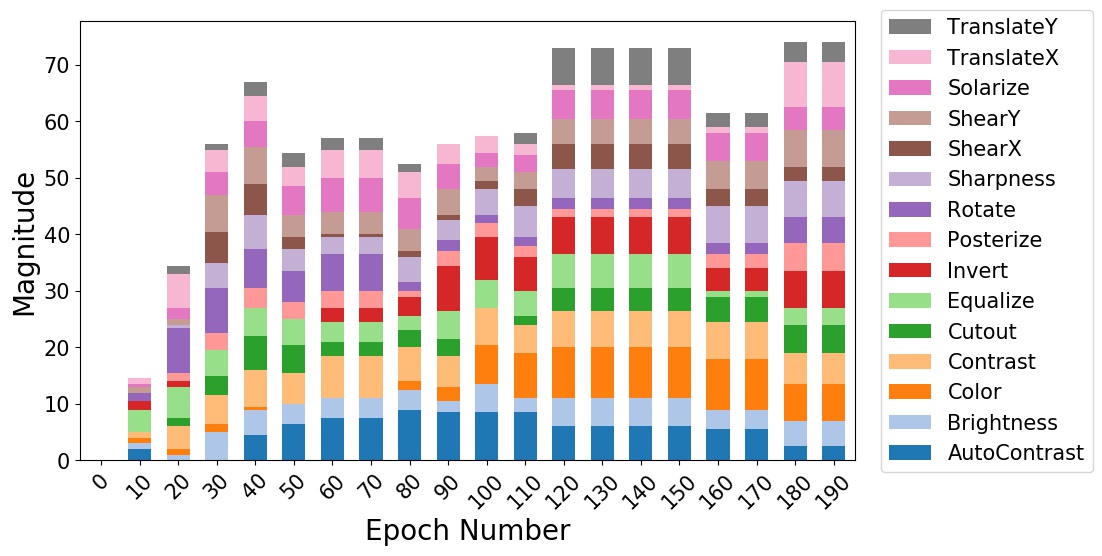}}\hfill
\caption{Evolution of magnitude and probability parameters in the learned schedules of teacher.
Each operation appears in the parameter list twice, and we take the mean values of the parameter.} 
\label{fig:vis_schedules_teacher}
\vspace{-6mm}
\end{figure}

\begin{figure}[!t]
\centering
\subfloat[Normalized plot of operation probability parameters over time for the student network $\mathcal{N}_S$.\label{fig:student_prob}]{\includegraphics[width=0.9\textwidth]{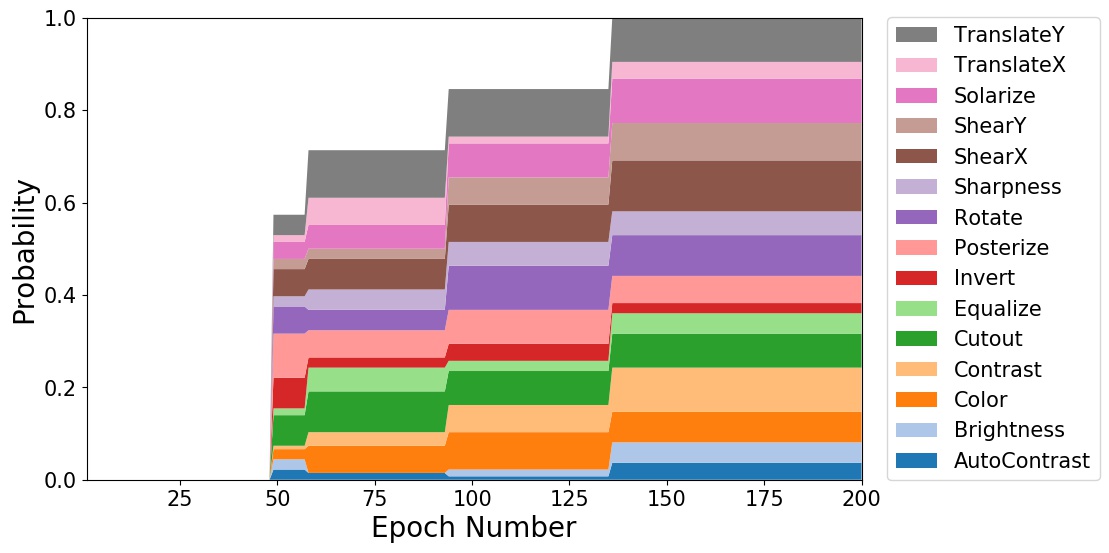}}\hfill
\subfloat[Operation magnitude parameters over time for the student network $\mathcal{N}_S$.\label{fig:student_mag}]{\includegraphics[width=0.9\textwidth]{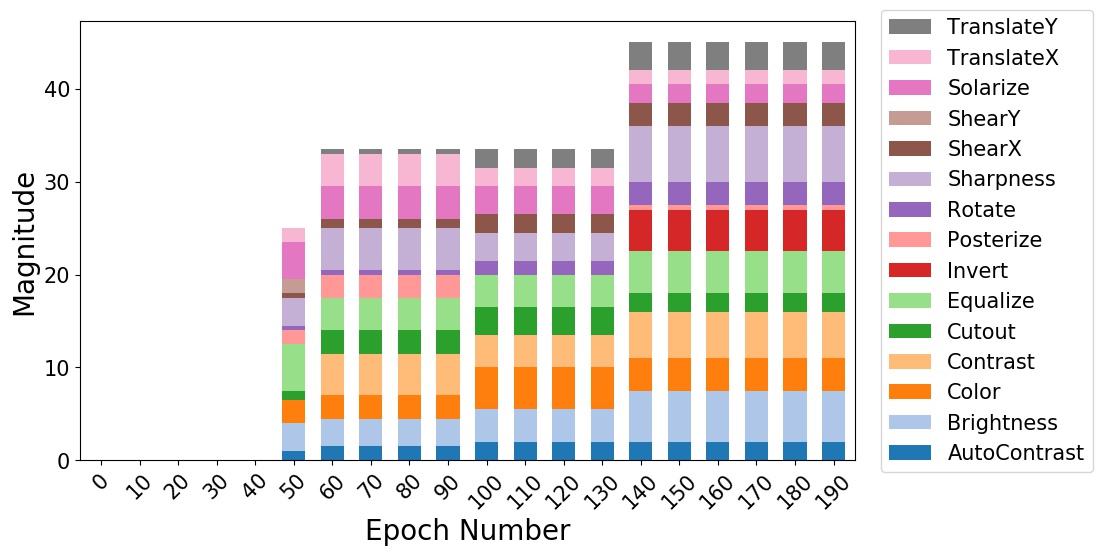}}\hfill
\caption{Evolution of magnitude and probability parameters in the learned schedules of student.} 
\label{fig:vis_schedules_student}
\vspace{-6mm}
\end{figure}

To analyze the difference on the discovered schedules between $\mathcal{N}_T$ (i.e., full-precision ResNet18) and $\mathcal{N}_S$ (i.e., 4-bit ResNet18), we report their augmented schedules quantitatively in terms of normalized probability and magnitude on CIFAR-100. Fig. \ref{fig:vis_schedules_teacher} shows the augmented schedules for teacher while Fig. \ref{fig:vis_schedules_student} shows the student. 
We normalize the probability of each epoch by dividing the maximal summation of probabilities for all operations across all epochs.

It can be seen that the discovered schedules $\mathcal{A}_S$ for $\mathcal{N}_S$ is quite different from $\mathcal{A}_T$ for $\mathcal{N}_T$. 
In particular, for $\mathcal{A}_T$, there is an emphasis on Brightness, Posterize, Rotate, Sharpness and TranslateY, while $\mathcal{A}_S$ cares more about Contrast, ShearX and TranslateY. 
Furthermore, we observe that the probability and magnitude increase as the epoch evolves. 
For $\mathcal{A}_S$, in the beginning, KD plays a more important role, and there is no augmentation operation before about epoch 50. 
As the training continues, the augmentation policies become more important. 
One possible reason is that, for low-precision networks, KD methods can provide rich training signals such that data augmentation does not help in the early training phases. 

Furthermore, we observe that, compared to $\mathcal{A}_T$, the schedules for student $\mathcal{A}_S$ evolves more smoothly in the sense that the policy updating frequency is lower. 
For example, the probability and magnitude values change about every 40 epochs for student, while the policies for teacher update about every 15 epochs. 
One possible reason might be that for the low-precision $\mathcal{N}_S$, KD methods make the training process more smooth and it is not necessary to change the augmentation policies too frequently. 
This is consistent with the observations shown in Tab. \ref{tab:baseline_comp} that KD can already provide useful training signals. 
It should be also noted that the teacher policies $\mathcal{A}_T$ become more smooth in the later stages than those in the early stages, and the learning of the student policies $\mathcal{A}_S$ can be seen as an extension and modification to $\mathcal{A}_T$, though in an indirect way by the KD signals. 
More importantly, this validates our assumption that $\mathcal{N}_S$ has her own optimal augmentation schedule $\mathcal{A}_S$ that is different from $\mathcal{A}_T$.







\subsection{Comparisons on Full-Precision Networks}
\label{sec:full_precision_experiment}
This subsection aims to verify the effectiveness of our proposed methods on more conventional KD settings where both $\mathcal{N}_T$ and $\mathcal{N}_S$ are full-precision networks. In particular, $\mathcal{N}_S$ is a shallow or a narrow network. We conduct experiments on various network achitectures, including Wide Residual Network (WRN) \cite{BMVC2016_87}, PyramidNet \cite{DPRN} and PreResNet. In particular, we use WRN-28-10 (we use the standard notation WRN-$d$-$k$ to refer to a wide residual network with depth $d$ and width multiplier $k$), PreResNet164 and PyramidNet-200-240 (PyramidNet-$d$-$k$ refers to a PyramidNet network with depth $d$ and widening factor $k$) as the teacher networks while WRN-16-2, PreResNet56, PreResNet44 and PreResNet32 as the student networks. 

\begin{table}[htb]
\centering
\caption{Accuracy on CIFAR-100 with full-precision under different settings for student network $\mathcal{N}_S$ and teacher network $\mathcal{N}_T$.
\textbf{Vanilla} refers to training a full-precision student network from scratch. \textbf{After Stage-$\alpha$} refers to using learned schedules discovered by PBA to re-train $\mathcal{N}_S$ as described in Section \ref{sec:stage_one}.
\textbf{only II-KD} refers to training $\mathcal{N}_S$ using II-KD but without the learned data augmentation.
\textbf{After Stage-$\beta$} refers to training $\mathcal{N}_S$ using II-KD and the learned data augmentation. The accuracies of teachers on CIFAR-100: WRN-28-10 has 80.73\%, ResNet164 has 72.24\% and PyramidNet-200-240 has 84.43\%. }

    \begin{tabular}{c|c|c|c|c|c}
    \hline
    Teacher & Student & Vanilla & After Stage-$\alpha$ & Only II-KD & After Stage-$\beta$ \\ \hline
    WRN-28-10 & WRN-16-2 & 72.68 & 73.79 & 74.41 &\cellcolor{JieGreen} 76.19 \\ \hline
    WRN-28-10 & WRN-16-4 & 77.28 & 78.01 & 79.30 &\cellcolor{JieGreen} 80.59 \\ \hline
    WRN-28-10 & PreResNet56 & 71.98 & 73.45 & 73.04 &\cellcolor{JieGreen} 74.91 \\ \hline
    PreResNet164 & PreResNet32 & 70.28 & 71.68 & 70.53 &\cellcolor{JieGreen} 72.14 \\ \hline
    PyramidNet-200-240 & PreResNet44 & 71.55 & 73.31 & 73.00 &\cellcolor{JieGreen} 73.71 \\ \hline
    PyramidNet-200-240 & WRN-16-2 & 72.68 & 73.79 & 74.64 &\cellcolor{JieGreen} 75.03 \\ \hline
    \end{tabular}
    \vspace{-4mm}
    \label{tab:32bit_comp}
\end{table}

Tab. \ref{tab:32bit_comp} reports the accuracy on CIFAR-100 under different network settings of teachers and students. 
We observe that the improvement of augmentation with KD is significant increase of about 3\% as compared to the vanilla baseline training. 
It shows that the discovered augmentation schedules further boosts the performance of the shallow $\mathcal{N}_S$ based on II-KD. 
In other words, our proposed method also works well on full-precision training tasks. Specifically, compared with only using KD, the learned schedules help to improve the performance by about 1.5\%. 

Furthermore, it shows that when the accuracy gap between the teacher and student is large, the teacher can guide the training of the student better, which is consistent with previous KD literature \cite{Mishra2018,Mirzadeh2019}. 
For example, considering distilling knowledge into the same student WRN-16-2, PyramidNet-200-240 (with accuracy 84.43\%) does a better job than WRN-28-10 (with accuracy 80.73\%). 
However, when combining KD and data augmentation, the PyramidNet-200-240 behaves worse than the WRN-28-10 as a teacher.
It seems that the improvement brought by the augmentation operations is more obvious when the teacher and student have similar network architectures.

\section{Conclusion}
Previous literature on KD focuses on exploring the knowledge representation and the strategies for distillation. However, both the teacher and student learn from the same training data without adapting the different learning capabilities. 
To address this issue, we propose customizing distinct agents to automatically augment the training data for the teacher and student, respectively. 
We extensively study the effectiveness of combining data augmentation and knowledge distillation. 
We also propose a simple feature-based KD variant that incorporates both intra- and inter-relationships within and across feature maps. 
We observe that the student can learn better from the teacher with the proposed approach, both in the challenging low-precision scenarios and with conventional full-precision networks. Furthermore, the teacher and student have their own optimal epoch-based augmentation schedules.

\clearpage
%
%
\bibliographystyle{splncs04}
\bibliography{aug_quantization}
\end{document}